\documentclass[11pt]{article}

\usepackage{arabtex}
\usepackage{utf8}
\setcode{utf8}

\usepackage{multirow}
\usepackage{amsmath}

\usepackage{acl}

\usepackage{subcaption}

\usepackage{times}
\usepackage{latexsym}

\usepackage[T1]{fontenc}
\usepackage[utf8]{inputenc}

\usepackage{microtype}

\usepackage{inconsolata}

\usepackage{graphicx}

\title{CANDLE: CTC-based Arabic Noisy-character Deduplication using a Lightweight Encoder}

\author{Faris Alasmary\textsuperscript{1} \quad Taif Nono\textsuperscript{1} \quad Orjuwan Zaafarani\textsuperscript{2}\thanks{\; Work done prior to joining SDAIA - NCAI.} \quad Kholood Al Tabash\textsuperscript{1}\\
  \textbf{Ahmad Ghannam\textsuperscript{1} \quad Anas Salamah\textsuperscript{1} \quad Shouq Sadah \textsuperscript{1} \quad Lahouari Ghouti\textsuperscript{1}} \\
  \textsuperscript{1}Abjad Ltd. \\
  \textsuperscript{2}SDAIA - NCAI \\
  \texttt{\{falasmary,tnono,kaltabash,aghannam,asalamah,ssadah,lghouti\}@abjad.com.sa} \\
  \texttt{o.zaafarani@sdaia.gov.sa}}

\begin{document}
\maketitle
\begin{abstract}
Handling repeated characters in text can be tricky, since they can represent either the correct spelling of a word or informal character elongation often seen in social media posts.
We present CANDLE, a lightweight system for character-level Arabic noise deduplication that addresses this challenge without relying on handcrafted rules, dictionaries, or morphological analyzers. At the heart of CANDLE is a novel application of Connectionist Temporal Classification (CTC) to this task, a formulation not previously explored for character deduplication, which frames normalization as a sequence alignment problem over a character-based encoder. Evaluated on three benchmarks spanning clean newspaper, manually curated ambiguous cases, and real-world social media text, the CTC model achieves a Sentence Error Rate (SER) as low as $5.37\%$ and consistently outperforms a classification-based baseline by a large margin. To reduce inference overhead, we distill the 6-layer CTC model into a 2-layer student, achieving a $3\times$ depth reduction with minimal performance degradation. Beyond deduplication accuracy, normalization yields a practical downstream benefit: a relative reduction in tokenizer fertility of up to $12.8\%$ across a diverse set of Arabic LLM tokenizers, directly lowering inference costs and improving context window utilization. We release all code and models publicly to support reproducibility and advance future research\footnote{\url{https://github.com/abjadai/candle}}.
\end{abstract}

\section{Introduction}

Text normalization is an important preprocessing step in Natural Language Processing (NLP), where the goal is to reduce variation in written text without changing its meaning.
One such normalization step is the removal of invalid consecutive character repetitions,  that is, repeated characters that do not contribute to the meaning of the word.
This issue is common across different languages, particularly in informal text on social media and online platforms, where writers frequently repeat characters within a word to express emphasis or emotion.
For example, an English speaker might write \textit{"sooooo good"} instead of \textit{"so good"}, or an Arabic speaker might write \textbf{\<مرررحبا>} instead of the correct form \textbf{\<مرحبا>}.
These repetitions have no grammatical or semantic value and can therefore be safely reduced to their standard single-character form.
However, a key challenge is telling apart this kind of noisy repetition from character sequences that are linguistically correct.
In Arabic, some repeated character patterns are not informal elongations but are actually valid parts of the word, and changing them would change the meaning.
A good example of this is the difference between \textbf{\<المملكة>} (\textit{al-mamlaka}, "the kingdom") and \textbf{\<الملكة>} (\textit{al-malika}, "the queen"), where the extra \textbf{\<م>} in \textbf{\<المملكة>} is not noise but an essential part of the word. Removing it would silently change the meaning of the text, as shown in the following examples:

\begin{center}
1. \<المملكة الأردنية الهاشمية>

    \begingroup
        \footnotesize
            Translation: The Hashemite Kingdom of Jordan. 
    \endgroup

2. \<الملكة الأردنية رانيا>

    \begingroup
        \footnotesize
            Translation: Queen Rania of Jordan.
    \endgroup
\end{center}

\noindent In the first example, \textbf{\<المملكة>} correctly means "\textit{kingdom}", while in the second, \textbf{\<الملكة>} means "\textit{queen}",  a reading that is further supported by the proper name \textbf{\<رانيا>} (\textit{Rania}).
Collapsing the repeated \textbf{\<م>} in the first example without considering the context would produce the incorrect phrase "\textit{The Hashemite Queen of Jordan}".
A good normalization system must therefore be able to tell the difference between the two cases, removing noisy repetitions while keeping the meaningful ones. In terms of how often valid repetitions occur, they rarely exceed two consecutive characters in Arabic, and three consecutive repetitions are very uncommon.
The theoretical maximum, however, is five, as shown by the classical Arabic example \<ما رأينا كُكَكًاَ كَكُكَكِكُم> (literally \<ما رأينا مراكبَ كباراً مثل مراكبكم>) \cite{ahmed1987subhalasha}, which means "\textit{We have never seen ships as large as yours}".
Any repetition beyond that can safely be treated as noise.

This paper presents a model for normalizing character-level repetitions in Arabic text. The model detects and collapses redundant consecutive character repetitions while keeping sequences that are linguistically valid.
Since valid repetitions in Arabic rarely exceed two consecutive characters, the model uses a maximum threshold of two, treating anything beyond that as noise.
The model works directly on character strings, without the need for phonological or morphological analysis, making it simple and efficient.

To formalize the notation used throughout this paper, we define two core reduction operations. The first, $reduce\_to\_1()$, collapses any sequence of identical consecutive characters down to a single instance. For example, "\<الله>" becomes "\<اله>" (reducing "\<لل>" to "\<ل>"), and "\textit{good}" becomes "\textit{god}". The second, $reduce\_to\_2()$, first applies $reduce\_to\_1()$ and then duplicates each character so that every character appears exactly twice. For example, "\<الله>" becomes "\<االلهه>", and "\textit{good}" becomes "\textit{ggoodd}". Notice that "\<ل>" and "\textit{o}" do not get duplicated. We refer to the output of $reduce\_to\_1()$ applied to a word as its \textit{reduced form}.

\section{Related Work}

Repeated-letter normalization is a key component of noisy text processing, particularly in social media, where character elongation is commonly used for emphasis or informal expression. Existing approaches can be broadly grouped into four categories: rule-based reduction methods, spell-checking and lexical similarity methods, corpus- and context-based methods, and neural sequence-to-sequence (seq2seq) models \cite{sutskever2014sequence}. This progression from rule-based to data-driven methods reflects a continuous effort to balance efficiency, robustness, and reliance on linguistic resources. However, significant challenges remain, especially when handling highly irregular or unseen repetition patterns.

\subsection{Rule-Based Reduction Approaches}

These methods normalize repeated characters by truncating any sequence of repeated letters to a fixed number of occurrences, typically one, two, or three. For example, \citet{rout2018model} limit characters to a maximum of three repetitions, while \citet{le2016sentiment} reduce repeated letters to a single occurrence (\textit{AABBBCC:ABC}) and repeated syllabic patterns to two occurrences (\textit{ABABABA:ABAB}) using handcrafted normalization rules. \citet{aliero2023systematic} reported that many social media normalization systems adopt similar methods.
Although simple and efficient, these methods do not account for lexical validity or contextual information, and may either incorrectly alter valid words with legitimate repetitions or fail to fully remove noise from heavily elongated forms.
This motivates the need for methods that can verify whether a candidate form is a valid word.

\subsection{Spell-Checking and Lexical Similarity Approaches}
These methods extend fixed-length reduction by generating candidate word forms after removing repetitions, then selecting the most likely candidate using spell-checking or lexical similarity measures. For example, \citet{sosamphan2016snet} applies regular expression-based reduction followed by spell-checking, while \citet{mansur2024normalization} use position-aware truncation with SymSpell validation. \citet{barik2019normalization} combine rule-based normalization with lexical similarity derived from word distribution, and \citet{desai2015normalization} shorten words so that they do not exceed two repetitions before the lexical matching with Levenshtein distance to find the nearest match.  \citet{hegazi2021preprocessing} normalize Arabic social media text by removing repetitions and matching the result against a predefined dictionary of non-standard forms.
While more robust than simple truncation, these methods still depend on linguistic resources and may struggle when multiple candidates have similar similarity scores, pointing to the need for broader contextual evidence to resolve such ambiguity.

\subsection{Corpus- and Context-based Approaches}
These methods rely on a dictionary of non-standard forms and use statistical evidence or surrounding context rather than a spell-checker alone.
\citet{darwish2012language} first check words against an MSA dictionary, then collapse repeated characters in unseen words in a way that is similar to our $reduce\_to\_1()$ operation and map them to the most frequent surface form found in a hash table that was constructed from a large tweet corpus.
\citet{demir2022graph} remove repetitions before computing edit distances and use graph-based contextual information to identify the intended word. While these methods can handle words not found in dictionaries, they may create ambiguity when multiple candidates have similar frequencies or contextual scores, and often struggle with unfamiliar slang, reduced forms, and dialect-specific words.
These limitations highlight the importance of  model-based approaches that can learn normalization patterns directly from text.

\subsection{Neural Sequence-to-Sequence Approaches}
 These methods treat repeated-character normalization as a data-driven task that maps noisy text directly to normalized forms without handcrafted rules or dictionaries. \citet{lourentzou2019adapting} propose a hybrid Seq2Seq model that combines word-level detection with a character-level denoising network trained on synthetically elongated words. \citet{partanen2019dialect} use LSTM- and Transformer-based neural machine translation models trained on parallel dialect–standard data to implicitly learn normalization of elongated forms.

\section{Dataset Preparation}
\subsection{Training Data}
To train our models, we used a subset of the 1.5 Billion Words Arabic Corpus \cite{abuelkhair20161Bwords}, combining the \textit{Alriyadh} and \textit{Alyaum} subsets and splitting the combined data into training and validation sets. We chose news articles as our training source since they are formally written, well-edited, and largely free of noisy character repetitions, making them a reliable source of clean Arabic text. During preparation, we retained only Arabic letters and removed all other characters. The dataset statistics are summarized in Table \ref{tab:train_data_summary}.

\begin{table}[ht!]
    %---------- Train set ----------
    % Chars: 2,832,074,618
    % Words: 480,608,963
    % Lines: 10,530,174
    % ------------------------------
    %------- Validation set --------
    % Chars: 2,807,959
    % Words: 476,642
    % Lines: 10,514
    \centering
    \small
    \begin{tabular}{c|ccc}
        \hline
        \textbf{Data Split} & \textbf{Chars} & \textbf{Words} & \textbf{Lines}\\
        \hline
        Train set & 2.83B & 480.61M & 10.53M \\
        Validation set & 2.81M & 476,642 & 10,514 \\
    \end{tabular}
    \caption{Training data summary}
    \label{tab:train_data_summary}
\end{table}

\subsection{Benchmark Datasets}
\subsubsection{NewsText Dataset}
One way to measure the errors introduced by the model is to use a clean corpus as a reference. In this setting, the model is expected to reproduce the input sentence exactly, preserving any valid character repetitions. To this end, we scraped a clean corpus from the Sabq\footnote{\url{https://sabq.org}} online newspaper. Following the same cleaning steps used for the training data, we retained only Arabic letters and removed all other characters and digits. We then identified all words in the dataset that contained more than two consecutive repeated characters and found $53$ such words. After manual inspection, most of them were found to contain spelling errors, which we corrected. Only six words were found to contain valid repetitions:
"\<بببتيد>",
"\<تتتوج>",
"\<تتتابع>",
"\<تتتبع>",
"\<ستتتبع>", and
"\<وتتتبع>".
The dataset statistics are summarized in Table \ref{tab:benchmark_datasets}.

\subsubsection{AmbigText Dataset}
We constructed a benchmark dataset to evaluate the model's ability to select the correct word in ambiguous cases, where different words share the same reduced form under $reduce\_to\_1()$ yet differ in meaning.
For example, the words
 "\<جدّ>" (\textit{grandfather}) and "\<جدّد>" (\textit{renewed}) both reduce to "\<جد>"  when $reduce\_to\_1()$ is applied.
 The dataset was manually collected and curated, and contains a total of $273$ sentences, covering $218$ Modern Standard Arabic (MSA) sentences and $55$ Saudi dialect sentences.
 These details are also shown in Table \ref{tab:benchmark_datasets}.

\subsubsection{WildSAText Dataset}
This dataset was collected to evaluate the models on real-world text.
We used a subset of the Saudi Forums Corpus (SFC) \cite{qarah2024saudibert} and applied the same cleaning steps as in the previous datasets, retaining only Arabic letters.
We then randomly sampled a subset of the SFC, excluding lines that contained words whose $reduce\_to\_1()$ output is a single character, such as 
"\<خخخخ>",
"\<هههه>",
or "\<آآآآ>",
as these do not represent meaningful words\footnote{These are common Arabic/Persian informal expressions used on social media to represent laughter and express surprise/pain/realization, respectively.}.
We also excluded sentences shorter than $12$ or longer than $256$ characters.
We used GPT 5.4 and Gemma4-it\footnote{\url{https://huggingface.co/google/gemma-4-31B-it}} to generate normalized reference outputs for the sampled data, and only kept samples where both models produced matching outputs.
After filtering, the dataset contains a total of $142,280$ sentences, of which $86,592$ contain more than two consecutive repetitions and $55,688$ contain two or fewer repetitions.

\begin{table}[ht!]
    %%%%%%%%% NewsText Benchmark %%%%%%%%%
    % Chars: 102,539,855
    % Words: 17,044,438
    % Lines: 488,571
    %%%%%%%%%%%%%%%%%%%%%%%%%%%%%%%%%%%%%%%%%%%%
    %%%%%%%%%%%%% WildSAText  Benchmark %%%%%%%%%%%%
    % Chars: 12,337,767
    % Words: 2,176,054
    % Lines: 142,280
    %%%%%%%%%%%%%%%%%%%%%%%%%%%%%%%%%%%%%%%%%%%%
    %%%%%%%%% AmbigText Benchmark %%%%%%%%%
    % Chars: 11,897
    % Words: 2,032
    % Lines: 273
    %%%%%%%%%%%%%%%%%%%%%%%%%%%%%%%%%%%%%%%%%%%%
    \centering
    \small
    \begin{tabular}{c|ccc}
        \hline
        \textbf{Benchmark Data} & \textbf{Chars} & \textbf{Words} & \textbf{Lines}\\
        \hline
        WildSAText  & 12.34M & 2.18M & 142,280 \\
        AmbigText & 11,897 & 2,032 & 273 \\
        NewsText & 102.54M & 17.04M & 488,571 \\
    \end{tabular}
    \caption{Details of the benchmark datasets}
    \label{tab:benchmark_datasets}
\end{table}

\begin{table*}[ht!]
    \centering
    \small
    \begin{tabular}{c|c|c|ccc}
        \hline
        \multirow{2}{*}{\textbf{Tokenizer}} & \multirow{2}{*}{\textbf{Raw Data}} & \multirow{2}{*}{\textbf{GroundTruth}} & \multicolumn{3}{c}{\textbf{Models}} \\
        \cline{4-6}
        & & & \textbf{CTC} & \textbf{CTC-Distilled} & \textbf{Classification} \\
        \hline
        JAIS 13B             & 1.5012 & 1.3534 (+9.85\%)  & 1.3510 (+10.01\%) & 1.3513 (+9.99\%) & 1.3535 (+9.84\%)  \\
        Gemma4 it 31B        & 1.9874 & 1.8180 (+8.52\%)  & 1.8147 (+8.69\%)  & 1.8148 (+8.69\%)  & 1.8111 (+8.87\%)  \\
        Llama 3.3 70B        & 2.3082 & 2.1353 (+7.49\%)  & 2.1318 (+7.64\%)  & 2.1318 (+7.64\%)  & 2.1209 (+8.12\%)  \\
        GPT-2                & 5.6239 & 5.3601 (+4.69\%)  & 5.3538 (+4.80\%)  & 5.3525 (+4.83\%)  & 5.3012 (+5.74\%)  \\
        Aranizer-SP-86k      & 1.4319 & 1.2820 (+10.47\%) & 1.2796 (+10.64\%) & 1.2802 (+10.60\%) & 1.2834 (+10.37\%) \\
        ALLaM 7B Instruct    & 1.4554 & 1.2712 (+12.65\%) & 1.2689 (+12.81\%) & 1.2696 (+12.76\%) & 1.2732 (+12.52\%) \\
        Command-R+           & 2.2404 & 2.0306 (+9.36\%)  & 2.0274 (+9.51\%)  & 2.0275 (+9.50\%)  & 2.0212 (+9.78\%)  \\
        Qwen 3.6 35B         & 2.0162 & 1.8096 (+10.25\%) & 1.8068 (+10.38\%) & 1.8069 (+10.38\%) & 1.8050 (+10.47\%) \\
        \hline
    \end{tabular}
    \caption{Fertility scores across models and tokenization methods. Values in parentheses indicate percentage improvement over Raw Data.}
    \label{tab:fertility_results}
\end{table*}

\section{Experiments}

Our models are initialized using a pretrained character-based BERT architecture \cite{alasmary2024catt}. Each model has 6 transformer layers with  $d_{model} = 512$ and $16$ attention heads, and was trained with an effective batch size of $ 1024 $ and a dropout rate of $ 10\% $.
We used the AdamW optimizer \cite{loshchilov2018decoupled} with a learning rate of $ 5 \times 10^{-5} $ and a weight decay of  $ 1 \times 10^{-2} $.
During data loading, lines exceeding the maximum sequence length are split into smaller segments with a stride of half the maximum sequence length. The minimum sentence length was set to $ 21 $ characters, which corresponds to approximately three words assuming an average word length of $6$ characters.

\subsection{Classification Approach}
In this approach, we apply $reduce\_to\_1()$ to the input sentence, then classify each character as either \textit{"don't duplicate"} or \textit{"duplicate"}.
Since $reduce\_to\_1()$ does not increase the sequence length of the input text, the maximum sequence length is set to $512$, matching the limit of the pretrained model.
Since most characters are not duplicated, the labels are heavily skewed toward \textit{"don't duplicate"}, resulting in a class imbalance problem.
To address this, we experimented with different class weights in the cross entropy loss and found that the best values are $0.3$ and $0.7$ for \textit{"don't duplicate"} and \textit{"duplicate"} respectively.
We started training by freezing all layers except the output layer, then gradually unfroze the transformer layers one by one. We found that the best model was the one with all transformer layers frozen and only the output layer trainable.

\subsection{CTC Alignment Approach}
We treat the problem as an alignment task where the model learns to recover the original correct text from a reduced input. We used the CTC loss \cite{graves2006connectionist} to find the alignment.
Since $reduce\_to\_2()$ always produces a sequence that is almost double the length of the original input text, the CTC assumption that 
the input sequence length, $ M $, is greater than that of the output sequence, $ N $, is always satisfied.
In our initial implementation, we generated random repetitions for each input character and used the original sentence as the target. However, this approach made learning difficult as the model failed to converge, likely due to the high variability introduced by the random repetitions.
To address this, we redesigned the input by applying $reduce\_to\_2()$, which ensures that every character is repeated exactly twice, providing the model with a consistent and predictable input structure. This simplification led to faster convergence within a few epochs.
Since $reduce\_to\_2()$ doubles the sequence length before feeding it to the model, the maximum input sequence length is set to $256$, so that the doubled sequence does not exceed the pretrained model's hard limit of $512$ \cite{alasmary2024catt}.
As the alignment task is harder than the classification task, freezing most layers was not sufficient. To fix this training issue, we unfroze the output layer and the last two transformer layers and trained until convergence. Then, we carried out another training step when the last three transformer layers are unfrozen. The initialization was based on the best checkpoint of the two-layer training step. Training was executed until convergence. In the final training step, we unfroze the entire transformer to train the model initialized from the best checkpoint of the three-layer phase. Like the previous steps, training was carried out until convergence.

\paragraph{Model Distillation.} To reduce the computational cost of the CTC alignment model while preserving its performance, we distilled the 6-layer teacher model into a 2-layer student model. The student model retains the same $d_{\text{model}} = 512$ and number of attention heads (16) as the teacher, ensuring that the representational capacity per layer is unchanged and only the depth is reduced.

Rather than initializing the student from scratch, we initialize its weights from the best checkpoint of the trained teacher model. Specifically, the first layer of the student is initialized with the weights of the teacher's first layer, and the second (final) layer of the student is initialized with the weights of the teacher's last layer \cite{shleifer2020pre,gandhi2023distil}. This strategy is motivated by the observation that the first layer captures low-level character representations while the last layer encodes the final decision boundary, making them the most transferable layers for a shallow student model.

The student is trained using a combination of soft and hard targets. The total loss is defined as:

\begin{equation}
    \mathcal{L} = \alpha \cdot \mathcal{L}_{\text{soft}} + (1 - \alpha) \cdot \mathcal{L}_{\text{hard}}
\end{equation}

\noindent where $\mathcal{L}_{\text{soft}}$ is the Kullback-Leibler (KL) divergence between the softened probability distributions of the teacher and student logits at temperature $T$, and $\mathcal{L}_{\text{hard}}$ is the standard CTC loss against the ground truth labels. We set $\alpha = 0.7$, giving more weight to the teacher's soft targets, and a distillation temperature of $T = 3.0$ to produce softer probability distributions that carry richer inter-class information \cite{hinton2015distilling}.

The student model is trained on the same data used to train the teacher model, with a minimum sequence length of $21$ and a maximum of $256$ tokens, and an effective batch size of $1024$. We use the AdamW optimizer with a learning rate of $5 \times 10^{-5}$ and a weight decay of $1 \times 10^{-2}$. Training continued until convergence and the best checkpoint was selected for evaluation.

\begin{table*}[h]
    \centering
    \small
    \begin{tabular}{c|c|c|c|c|c}
    \textbf{Subset}                    & \textbf{Samples}         & \textbf{Metric} & \textbf{Classification} & \textbf{CTC} & \textbf{Distilled-CTC} \\ \hline
    \multirow{3}{*}{\textbf{All}}      & \multirow{3}{*}{488,571} & \textbf{SER}    & 63.79                   & 5.37         & 6.31                   \\
                                       &                          & \textbf{WER}    & 3.85                    & 0.18         & 0.21                   \\
                                       &                          & \textbf{CER}    & 0.78                    & 0.04         & 0.04                   \\ \hline
    \multirow{3}{*}{\textbf{w/ dups}}  & \multirow{3}{*}{335,062} & \textbf{SER}    & 93.01                   & 7.83         & 9.20                   \\
                                       &                          & \textbf{WER}    & 4.83                    & 0.22         & 0.26                   \\
                                       &                          & \textbf{CER}    & 0.97                    & 0.04         & 0.05                   \\ \hline
    \multirow{3}{*}{\textbf{w/o dups}} & \multirow{3}{*}{153,509} & \textbf{SER}    & 0.00                    & 0.00         & 0.00                   \\
                                       &                          & \textbf{WER}    & 0.00                    & 0.00         & 0.00                   \\
                                       &                          & \textbf{CER}    & 0.00                    & 0.00         & 0.00                  
    \end{tabular}
    \caption{Results on NewsText (in \%)}
    \label{results_newstext}
\end{table*}

\begin{table*}[h]
    \centering
    \small
    % \usepackage{multirow}
    \begin{tabular}{c|c|c|c|c|c}
    \textbf{Subset}                    & \textbf{Samples}     & \textbf{Metric} & \textbf{Classification} & \textbf{CTC} & \textbf{Distilled-CTC} \\ \hline
    \multirow{3}{*}{\textbf{All}}      & \multirow{3}{*}{273} & \textbf{SER}    & 69.96                   & 23.08        & 25.27                  \\
                                       &                      & \textbf{WER}    & 13.63                   & 3.15         & 3.59                   \\
                                       &                      & \textbf{CER}    & 2.96                    & 0.68         & 0.79                   \\ \hline
    \multirow{3}{*}{\textbf{w/ dups}}  & \multirow{3}{*}{192} & \textbf{SER}    & 99.48                   & 32.81        & 35.94                  \\
                                       &                      & \textbf{WER}    & 18.86                   & 4.36         & 4.97                   \\
                                       &                      & \textbf{CER}    & 4.01                    & 0.93         & 1.07                   \\ \hline
    \multirow{3}{*}{\textbf{w/o dups}} & \multirow{3}{*}{81}  & \textbf{SER}    & 0.00                    & 0.00         & 0.00                   \\
                                       &                      & \textbf{WER}    & 0.00                    & 0.00         & 0.00                   \\
                                       &                      & \textbf{CER}    & 0.00                    & 0.00         & 0.00                  
    \end{tabular}
    \caption{Results on AmbigText (in \%)}
    \label{results_ambigtext}
\end{table*}

\begin{table*}[h]
    \centering
    \small
    \begin{tabular}{c|c|c|c|c|c}
    \textbf{Subset}                    & \textbf{Samples}         & \textbf{Metric} & \textbf{Classification} & \textbf{CTC} & \textbf{Distilled-CTC} \\ \hline
    \multirow{3}{*}{\textbf{All}}      & \multirow{3}{*}{142,280} & \textbf{SER}    & 53.93                   & 9.93         & 12.43                  \\
                                       &                          & \textbf{WER}    & 5.73                    & 0.73         & 0.91                   \\
                                       &                          & \textbf{CER}    & 1.31                    & 0.17         & 0.23                   \\ \hline
    \multirow{3}{*}{\textbf{w/ dups}}  & \multirow{3}{*}{86,592}  & \textbf{SER}    & 88.58                   & 15.73        & 17.33                  \\
                                       &                          & \textbf{WER}    & 7.83                    & 0.96         & 1.07                   \\
                                       &                          & \textbf{CER}    & 1.78                    & 0.22         & 0.25                   \\ \hline
    \multirow{3}{*}{\textbf{w/o dups}} & \multirow{3}{*}{55,688}  & \textbf{SER}    & 0.05                    & 0.90         & 4.81                   \\
                                       &                          & \textbf{WER}    & 0.00                    & 0.09         & 0.49                   \\
                                       &                          & \textbf{CER}    & 0.00                    & 0.03         & 0.18                  
    \end{tabular}
    \caption{Results on WildSAText (in \%)}
    \label{results_wildsatext}
\end{table*}

\section{Results}
Tables~\ref{results_newstext}-\ref{results_wildsatext} present the results of the Classification and CTC models across the three benchmark datasets. Overall, the CTC model consistently outperforms the Classification model by a large margin across all datasets and metrics.

On the clean NewsText dataset, the CTC model achieves a Sentence Error Rate (SER) of just $5.37\%$ overall, compared to $63.79\%$ for the Classification model.
The largest difference appears in sentences containing repeated characters, where the Classification model reaches a SER of $93.01\%$ while CTC keeps it at $7.83\%$.
Notably, both models produce zero errors on sentences without duplicate characters, confirming that the models do not alter an already-clean text.

On the more challenging AmbigText dataset, the input to the model is ambiguous in the sense that the reduced form of a word could either remain reduced or be restored to its original duplicated form, depending on the surrounding context. The CTC model outperforms Classification with an overall SER of $23.08\%$ versus $69.96\%$, suggesting that the alignment-based formulation gives the model a stronger ability to leverage contextual information when resolving such ambiguities.

On the WildSAText dataset, which reflects real-world noisy Arabic text, the CTC model achieves an overall SER of $9.93\%$ compared to $53.93\%$ for the Classification model. Even on sentences without duplicates, the CTC model remains reliable, with a SER of only $0.90\%$, demonstrating strong generalization to informal and dialectal Arabic.

The Distilled-CTC model is a 2-layer student compressed from the 6-layer CTC teacher, a $3\times$ reduction in model depth, and closely tracks the teacher across all three benchmarks.
On NewsText, the overall SER increases by less than $1$ percentage point from $5.37\%$ to $6.31\%$, with CER remaining identical at $0.04\%$.
On AmbigText, the distilled model achieves a SER of $25.27\%$, slightly higher than the teacher's $23.08\%$.
On WildSAText, the overall SER rises modestly from $9.93\%$ to $12.43\%$, with the largest relative gap on sentences without repeated characters, where SER increases from $0.90\%$ to $4.81\%$.
Overall, the performance degradation is small relative to the $3\times$ model depth reduction.

Table~\ref{tab:fertility_results} shows that normalization yields a practical downstream benefit beyond deduplication: reduced tokenization overhead. Across all tested tokenizers on the WildSAText dataset, both the CTC and Distilled-CTC models reduce fertility scores relative to raw input by $4.8\%$
to $12.8\%$, closely tracking the ground truth range of $4.7\%$ to $12.7\%$. The two models perform nearly identically on this measure, indicating that compressing the CTC model to two layers preserves its tokenization benefits. This reduction in tokenization can translate directly to lower LLM inference costs and more efficient use of context windows.

\section{Conclusion}
 
We presented CANDLE, a lightweight system for normalizing character-level repetitions in Arabic text. We introduced a novel application of CTC-based sequence alignment to this problem, to our knowledge the first of its kind for character deduplication, which consistently and substantially outperforms the classification baseline across all benchmarks without relying on rules, dictionaries, or morphological analyzers.
We further distilled the 6-layer CTC model into a 2-layer student with minimal performance degradation, yielding a $3\times$ depth reduction that makes CANDLE practical for production deployments. Normalization also reduces tokenizer fertility by up to $12.8\%$ across diverse Arabic LLM tokenizers, directly lowering inference costs.
In future work, we plan to integrate prefix beam search with an $n$-gram language model \cite{hannun2014first} as a replacement for the greedy decoder, and to further investigate a 12-layer character-based BERT model under the classification approach, which showed competitive preliminary results.

\section*{Limitations}

Despite its contributions to Arabic text normalization, this system has two key limitations.

\begin{itemize}
    \item \textbf{Pure Arabic input requirement}: The system assumes all input is purely Arabic, 
    with no support for diacritics, numerals, or special characters. 
    Removing such elements risks distorting sentence structure and compromising normalization quality. 
    For example, dropping the numeral from \<اشتريت 3 كتب> (I bought 3 books) yields the malformed \<اشتريت كتب> (I bought books), 
    whereas the correct approach is to convert numerals to their Arabic word equivalents 
    such as \<اشتريت ثلاثة كتب> (I bought three books) before processing. 
    Therefore, a dedicated preprocessing stage prior to the deduplication module is strongly recommended.

    \item \textbf{Bounded repetition handling}: The system assumes no character appears consecutively 
    more than twice in any word. As a result, valid Arabic words with three or more consecutive 
    identical characters may be incorrectly collapsed. Examples of such valid words include
    "\<تتتوج>" (gets crowned),
    "\<تتتابع>" (keeps following),
    "\<تتتالى>" (follows in succession), and
    "\<فتتت>" (crumbles).

\end{itemize}

\section*{Ethics Statement}

This work adheres to the ACL Code of Ethics. Our data is drawn from the
publicly available 1.5 Billion Words Arabic Corpus, the Saudi Forums Corpus,
and a publicly accessible online newspaper, all cited to their original
sources. During preprocessing we retain only Arabic characters and discard all
other content, so no author identities, usernames, or account-level metadata
are collected or released, and we make no attempt to profile or re-identify
individuals.

Reference outputs for the WildSAText benchmark were produced by large language
models and kept only when the models agreed. These references may reflect the
biases of the underlying models and are not a substitute for exhaustive human
annotation. In addition, our system is developed and evaluated on Modern
Standard Arabic and Saudi dialectal text, and its performance may degrade on
other dialects, domains, or mixed-script input.

Because normalization is a preprocessing step, errors such as collapsing a
meaningful repetition can propagate to downstream tasks, so we recommend
applying the system with appropriate validation rather than as an unchecked
transformation of user text. To support transparency and reproducibility, we
release our code and models and report dataset statistics and training details
in the paper.


\bibliography{custom}

% \appendix

% \section{Example Appendix}
% \label{sec:appendix}

% This is an appendix.

\end{document}